# Disentangling Granularity: An Implicit Inductive Bias in Factorized VAEs

Zihao Chen, Yu Xiang*, Wenyong Wang*

*University of Electronic Science and Technology of China, Chengdu, 611731, China*

**Abstract**

Despite the success in learning semantically meaningful, unsupervised disentangled representations, variational autoencoders (VAEs) and their variants face a fundamental theoretical challenge: substantial evidence indicates that unsupervised disentanglement is unattainable without implicit inductive bias, yet such bias remains elusive. In this work, we focus on exploring the implicit inductive bias that drive disentanglement in VAEs with factorization priors. By analyzing the total correlation in $\beta$-TCVAE, we uncover a crucial implicit inductive bias called disentangling granularity, which leads to the discovery of an interesting "V"-shaped optimal Evidence Lower Bound (ELBO) trajectory within the parameter space. This finding is validated through over 100K experiments using factorized VAEs and our newly proposed model, $\beta$-STCVAE. Notably, experimental results reveal that conventional factorized VAEs, constrained by fixed disentangling granularity, inherently tend to disentangle low-complexity feature. Whereas, appropriately tuning disentangling granularity, as enabled by $\beta$-STCVAE, broadens the range of disentangled representations, allowing for the disentanglement of high-complexity features. Our findings unveil that disentangling granularity as an implicit inductive bias in factorized VAEs influence both disentanglement performance and the inference of the ELBO, offering fresh insights into the interpretability and inherent biases of VAEs.

*Keyords:* Variational Autoencoder (VAE); Implicit Inductive Bias; Factorization Prior; Total Correlation

---

* These authors are co-corresponding author: Wenyong Wang (wangwy@uestc.edu.cn); Yu Xiang (jcxiang@uestc.edu.cn, ORCID: 0000-0001-9622-7661).

## 1. Introduction

In recent years, the latent representation theorem has significantly improved the architecture of deep learning and has been empirically proven in terms of unsupervised representation learning[1, 2, 3], speech recognition [4, 5, 6], face recognition [7, 8, 9, 10], target monitoring [11, 12, 13], and natural language processing[14, 15, 16].

In 2017, βVAE [17] was proposed to illustrate that a certain degree of penalty could be imposed on the KL term of VAE's ELBO to disentangle the independence features of the posterior approximate distribution. To better trade-off disentanglement and reconstruction, works [18, 19] represented by β-TCVAE [20] refined the disentangling constraint term to focus on the independent marginal distributions of latent variables. These methods generally regularized the factorization prior among the latent variables to achieve better disentangling performance[21, 22].

As research progressed, a potentially disruptive conclusion emerged. In 2019, the ICML best paper provided a theoretical proof that unsupervised entanglement and disentanglement models were statistically equivalent [23]. Specifically, in the absence of inductive bias from either the model or the dataset, factorized VAEs failed to consistently identify or converge to disentangled parameterized representations[24, 25]. This study revealed that VAEs might naturally have implicit inductive biases that promote disentanglement, such a finding subsequently confirmed by further research[26]. From the perspective of generative factors, research has shown that all unsupervised VAE-based disentangling model succeed by leveraging the same structural bias inherent in the data. The ground true generative factors align closely with the nonlinear principal components that VAEs naturally extract. This bias could be mitigated by introducing minor modifications to the local correlation structure of the input data[27]. From the perspective of optimization algorithms, studies have demonstrated that, within an SGD-optimized framework, disentangled models exhibit faster convergence than entangled ones. This suggests that SGD itself may act as a natural inductive bias, fundamentally influencing the convergence efficiency of unsupervised disentanglement models[28].

Although these preliminary studies underscored the crucial role of implicit inductive biases in enabling unsupervised models to achieve disentanglement, the implicit inductive biases inherent in the model's statistical prior remain largely unexplored[29], particularly the factorization prior in VAEs, which serves as the central assumption in the theoretical proof presented in the ICML Best Paper[23]. Despite this, scarcely any research has investigated how it affects disentanglement and inference in factorized VAEs. Exploring and understanding these inductive biases is crucial, as they may provide more principled explanations for VAEs[30, 31].

Motivated by this gap, we explore the implicit inductive biases that govern disentanglement in factorization prior. Specifically, we begin by analyzing the total correlation term that determines the factorized prior in $β$-TCVAE and identify a crucial inductive bias—disentangling granularity, which regulates the model's tendency to disentangle features of varying complexity and influences the ELBO inference. By decomposing the total correlation term, we introduce a novel VAE model that allows for explicit tuning of disentangling granularity, without introducing additional structural complexity. Through the training of over 100K factorized VAEs, we identify a distinctive "V"-shaped optimal ELBO trajectory in the parameter space defined by disentangling granularity and network capacity. This phenomenon suggests that the fixed disentangling granularity imposed by the factorization prior in conventional VAEs directs them to disentangle low-complexity features while restricting their ability to disentangle high-complexity features. Reasonable tuning of disentanglement granularity could enhance both representation learning quality (ELBO) and disentanglement performance. These findings provide valuable insights into the correlation between disentanglement and the effectiveness of representation learning in VAEs.

The major contributions of this paper are as follows:

**Contributions**

- We uncover an implicit inductive bias, the disentanglement granularity, that governs the expression of disentanglement in factorized VAEs, and show that the factorized prior constrained by total correlations in $β$-TCVAE cause the disentanglement granularity to be fixed, preventing it from achieving better disentanglement performance and inferring higher ELBO.

- We propose a novel total correlation decomposition method, which explains the cause of VAE disentangling performance by gradually breaking down the total correlation. Using this decomposition method, we present a modified VAEs——$\beta$–STCVAE that allows for tuning disentangling granularity to flexibly control disentangling tendency.

- Experiments over 100K VAEs instances on different datasets suggest that：

(1) the proposed model is superior in both representation learning and disentangling capabilities.

(2) a "V"-shaped optimal ELBO trajectory pattern between model parameter capacity and the disentangling granularity is observed across all datasets, which indicates that by tuning an appropriate disentangling granularity, the new model consistently achieves better representation learning ability compared to factorized VAEs without additional structural complexity.

**2. Total Correlation Decomposition**

Table 1 provides explanations of the mathematical notations used in the subsequent chapters for ease of comprehension.

The $\beta$-TCVAE [20], as a representative of factorized unsupervised disentangled VAEs, was proposed in 2018. It decomposes the KL divergence term $\mathbb{E}_{p(x)}[\mathcal{D}_{KL}(q(z|x)||p(z))]$ in VAEs and obtains the following objectives:

$$\mathbb{E}_{p(x)}[\mathcal{D}_{KL}(q(z|x)||p(z))] = \mathcal{D}_{KL}(q(z,x_n)||q(z)p(x_n)) + \mathcal{D}_{KL}(q(z)||\prod_i q(z_i))$$
$$+ \sum_i \mathcal{D}_{KL}(q(z_i)||p(z_i)).$$

Based on this, the loss function of $\beta$-TCVAE can be written as:

$$L_{\beta-TCVAE} \coloneqq \mathbb{E}_{q(Z|X)p(x)}[log p(x|z)] - I(z\;;\;x) - \beta[\mathcal{TC}(z)] - \sum_i \mathcal{D}_{KL}(q(z_i)||p(z_i))$$
(1)

It is believed that the disentangling capability of $\beta$-TCVAE mainly comes from two aspects in (1): 1). The mutual information between latent variables and data variables $I(x;z)$; 2). The independence among latent variables $\mathcal{TC}(z)$. Here we focus on the total correlation item because the disentangling function of the $I(x;z)$ is still controversial [20].

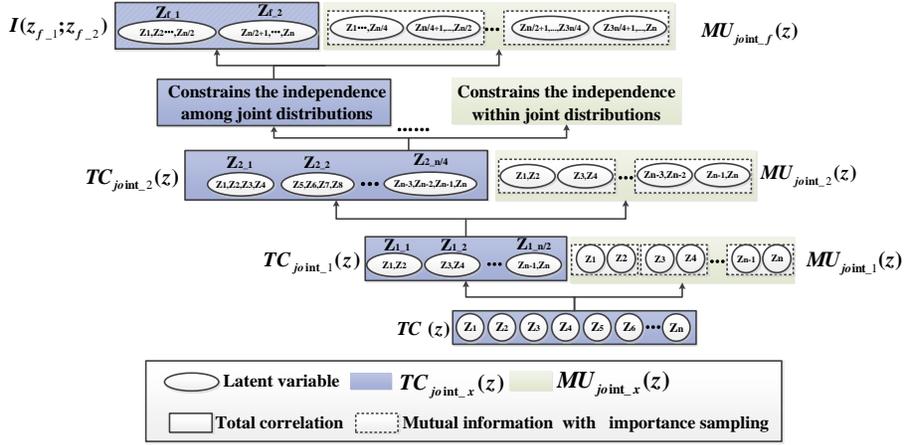

Fig 1. Bottom-up is the iterative decomposition process of $\mathcal{TC}$. The joint distribution total correlation, denoted as $\mathcal{TC}_{\text{joint}}(z)$. The importance sampling of mutual information between the marginal distributions of the internal, denoted as $\mathcal{MU}_{\text{joint}}(z)$. The total correlation is iteratively decomposed into the sum of $\mathcal{TC}_{\text{joint}}(z)$ and $\mathcal{MU}_{\text{joint}}(z)$.

In this paper, we propose a method to iteratively decompose the total correlation $\mathcal{TC}(z)$, whose complete decomposition process could be found in the appendix A.1. The following is a brief introduction to the novel decomposition method.

The Fig. 1 shows the visualization and intuitive understanding for decomposition process. Assuming that latent space $z$ consists of $n$ variables, then there is:

$$\mathcal{TC}(z) = \mathcal{D}_{KL}(q(z) \| \prod_k^n q(z_k)) =$$
$$\begin{cases} \sum_{cp(i,j)} I(z_i\,;\,z_j)|_{q(z)=q(z_i,z_j)\cdot q(z_{-i,j})} + \mathcal{D}_{KL}(q(z)\|\prod_{cp(i,j)} q(z_i,z_j)) &, e^* \\ \sum_{cp(i,j)} I(z_i\,;\,z_j)|_{q(z)=q(z_i,z_j)\cdot q(z_{-i,j})} + \mathcal{D}_{KL}(q(z)\|\prod_{cp(i,j)} q(z_i,z_j)\cdot q(z_r)) &, o^* \end{cases}$$
(2)

where $cp(i,j)$ denotes a mechanism for selecting a pair of latent variables $z_i, z_j$ without replacement.

The first term $\sum_{cp(i,j)} I(z_i\,;\,z_j)|_{q(z)=q(z_i,z_j)\cdot q(z_{-i,j})}$ is called "mutual information summation", which denotes the accumulation of mutual information combinations of all pairs $z_i, z_j$ of latent variables selected by $cp(i,j)$.

The second term in Eq. (2) is called the "joint distributions total correlation". During derivation, we discovered that the target distribution of the importance sampling in the first term $I(z_i\ ;\ z_j)$ satisfy $q(z_i, z_j)$ and $q(z_{-i,j})$ are independent of each other.

Then, we have $\mathcal{TC}_{joint\_1}(z)$:

$$\mathcal{TC}_{joint\_1}(z) := \begin{cases} \mathcal{D}_{KL}(q(z) || \prod_{cp(i,j)} q(z_i, z_j)) & , e^* \\ \mathcal{D}_{KL}(q(z) || \prod_{cp(i,j)} q(z_i, z_j) \cdot q(z_r)) & , o^* \end{cases} \quad (3)$$

Replacing $\sum_{cp(i,j)} I(z_i\ ;\ z_j)|_{q(z)=q(z_i,z_j)\cdot q(z_{-i,j})}$ with $\mathcal{MU}_{joint\_1}(z)$, there is $\mathcal{TC}_{joint\_1}(z) = \mathcal{TC}(z) - \mathcal{MU}_{joint\_1}(z)$.

From Eq. (2), we know $\mathcal{MU}_{joint\_1}(z)$, defined as the accumulation of mutual information, is always positive. Thus, there is always $\mathcal{TC}_{joint\_1}(z) \leq \mathcal{TC}(z)$. This inequality brings us two interesting inferences:

**Inference 1.** The total correlation of a joint distribution cannot exceed the sum of the total correlations of its marginal distributions.

**Inference 2.** We can split and refine the independent constraint granularity of $\mathcal{TC}(z)$: $\mathcal{MU}_{joint}(z)$ constrains the independence of marginal distributions within joint distributions, while $\mathcal{TC}_{joint}(z)$ constrains the independence among joint distributions.

After iterating the decomposition to its final form, we obtain the following equation:

$$\mathcal{TC}(z) = \sum_{b=1}^{f} \mathcal{MU}_{joint\_b}(z) + I(z_{f\_1}\ ;\ z_{f\_2}), \quad (4)$$

where $f = int(log_2 n) - 1$ denotes total decomposition iteration. If $f < 1$, then $z_{f\_i} = z_i$, and $n$ is the total latent variables number of $z$, and $int(\cdot)$ is the rounded-up integer.

## 3. Proposed Method

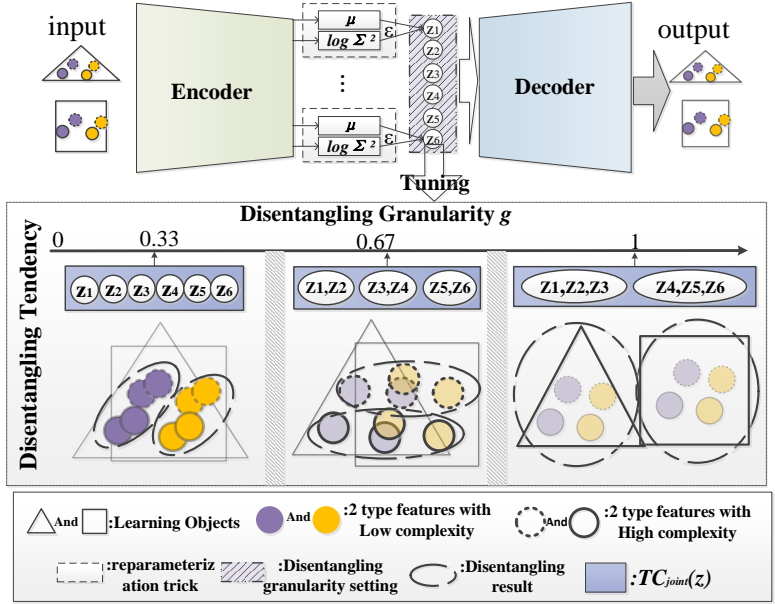

Fig 2: By tuning the disentangling granularity, we can control the disentangling expression of VAEs under different feature complexities. We use transparency to express the degree of disentangling attention that the model pays to the data features. The lower the transparency of the feature is, the more focused the disentangling attention from the model will be. As the disentangling granularity becomes smaller, the model focuses more on disentangling subtle features with lower complexity, while the disentangling granularity becomes larger, the model pays more attention to disentangle features with higher complexity.

The learned data features $x \in \mathbb{R}^N$ can be conceptually divided into two categories: conditionally independent features $v \in \mathbb{R}^K$ and conditionally dependent features $w \in \mathbb{R}^H$ [17]. The true data distribution can be approximated by a parametric neural network $p(x|v,w) = net(v,w)$, which is used to model the inferred posterior distribution $q(z|x)$.

From an intuitive perspective, the total parameter capacity ($\mathcal{C}_z$) of the network is similarly divided into two parts: one allocated to independent features ($\mathcal{C}_v$) and the other to dependent features ($\mathcal{C}_w$). Effective utilization of parameter capacity occurs while its allocation aligns closely with the complexity of these feature groups. Otherwise, a mismatch can hinder the model's learning ability.

However, without knowing the complexity of feature distribution, the disentangling method matching prior distribution among latent variables is gained during training, and the fixed factorized prior distribution is not necessarily
7

consistent with the real distribution of $x$, which poses a challenge to the disentangling effects of factorized VAEs.

*3.1. Disentangling Granularity and Novel Model $\beta$-STCVAE*

Inference 2 in the previous section 2 provides us with the motivation to fine-tune the granularity of independence constraints in the latent joint distributions serves as an implicit inductive bias control mechanism, enabling the model to adapt its disentanglement behavior to features of varying complexity.

Intuitively, removing the independence constraint on $\mathcal{MU}_{joint}(z)$ within the joint distribution is essential to ensure that a complete feature representation can be captured within this joint distribution. As the number of latent variables composing the joint distribution in $\mathcal{TC}_{joint}(z)$ increases, the model will be encouraged to learn independent features with high-complexity distribution.

After $b$ iterations of decomposition, and removing the constraints of $\mathcal{MU}_{joint\_1}(z) \sim \mathcal{MU}_{joint\_b-1}(z)$, then, there is:

$$\mathcal{TC}_{joint\_b}(z) := \begin{cases} \mathcal{D}_{KL}(q(z) || \prod_{cp(b-1\_i,b-1\_j)} q(z_{b-1\_i}, z_{b-1\_j})) & , e^* \\ \mathcal{D}_{KL}(q(z) || \prod_{cp(b-1\_i,b-1\_j)} q(z_{b-1\_i}, z_{b-1\_j}) \cdot q(z_r)) & , o^* \end{cases}$$

(5)

where $z_{b-1\_i}$ denotes the $i$th latent variable after $b-1$ decomposition iteration. If $b \leq 1$, then $z_{b-1\_i} = z_i$, and $z_{b-1\_j} = z_j$.

**Disentangling Granularity:**

With the above foundation, we can specifically define the inductive bias as follows.

We propose to control the depth of $\mathcal{TC}_{joint\_b}(z)$ decomposition and the disentangling tendency of different complexity features by defining a fine-tunable normalized implicit inductive bias $\boldsymbol{g}$:

$$S = \{\hat{b} \in \mathbb{N}^+ | \; n \equiv 0 \; (mod \; \hat{b})\}$$
$$m = max(S)$$
$$\boldsymbol{g} = \hat{b}/m$$
$$\text{s.t.} \;\; 1/m \leq \boldsymbol{g} \leq 1 \tag{6}$$

Where $n \equiv 0 \pmod{\hat{b}}$ denotes iteration round $\hat{b}$ is a divisor of $n$, named the grouping factor, which controls the gradual decomposition iterations of $\mathcal{TC}(z)$. $\boldsymbol{g}$ denotes the disentangling granularity, controlling the model's tendency to disentangle features with different complexity. Then we define $\mathcal{TC}_{joint\_\hat{b}}(z) := \mathcal{D}_{KL}(q(z) \| \prod_{j=1}^{n/\hat{b}} q(z_{\hat{b}(j-1)+1}, z_{\hat{b}(j-1)+2}, \ldots, z_{\hat{b}j}))$, as a variant of $\mathcal{TC}_{joint\_b}(z)$ in (5) implemented by a certain $cp(\cdot,\cdot)$.

Consequently, we propose modifying the $\mathcal{TC}(z)$ term in the $\beta$-TCVAE loss (1) by substituting it with $\mathcal{TC}_{joint\_\hat{b}}(z)$, thereby defining the loss function for the new model $\beta$-STCVAE:

$$L_{\beta-STCVAE} := \mathbb{E}_{q(z|x)p(x)}[log p(x|z)] - I(z;x) - \beta[\mathcal{TC}_{joint\_\hat{b}}(z)] - \sum_l \mathcal{D}_{KL}(q(z_l)\|p(z_l)),$$
(7)

Here, we give an example of a specific disentangling granularity setting process: while the $n$ is 12, the available grouping factors' set $S$ is {1, 2, 3, 4, 6}. We normalize grouping factors to get a unified view under different numbers of latent variables: Take the largest factor $m$ in the available grouping factors list as the normalization constant, and all the selected grouping factors $\hat{b}$ are recorded as $\hat{b}/m$, which becomes a coefficient less than or equal to 1 and greater than $1/m$, called the disentangling granularity $\boldsymbol{g}$. Refer to Fig. 2 for $\beta$-STCVAE structure.

We can qualitatively analyze the impact of disentangling granularity on the newly proposed model $\beta$-STCVAE. As can be seen from Fig. 2, the independence constraint between latent variables of the model weakens as $\boldsymbol{g}$ increases. The joint distribution of latent variables, which loses its independence constraint, has the opportunity to form a complex posterior distribution, learning to accommodate feature with higher complexity. In contrast, the smaller $\boldsymbol{g}$ is, the stronger the independent constraints of the model prior is. While $\boldsymbol{g} = 1/m$, The loss function (7) is equivalent to the loss function (1), i.e. $\beta$-STCVAE degenerate into $\beta$-TCVAE, and the prior distribution gradually becomes the multi-dimensional Gaussian distribution with independent marginal distributions. At this point, the model is more suitable for disentangling feature with lower complexity. The disentangling granularity $\boldsymbol{g}$ should be tuned based on the model application scenario, severing as an inductive bias for unsupervised disentangled representation learning.

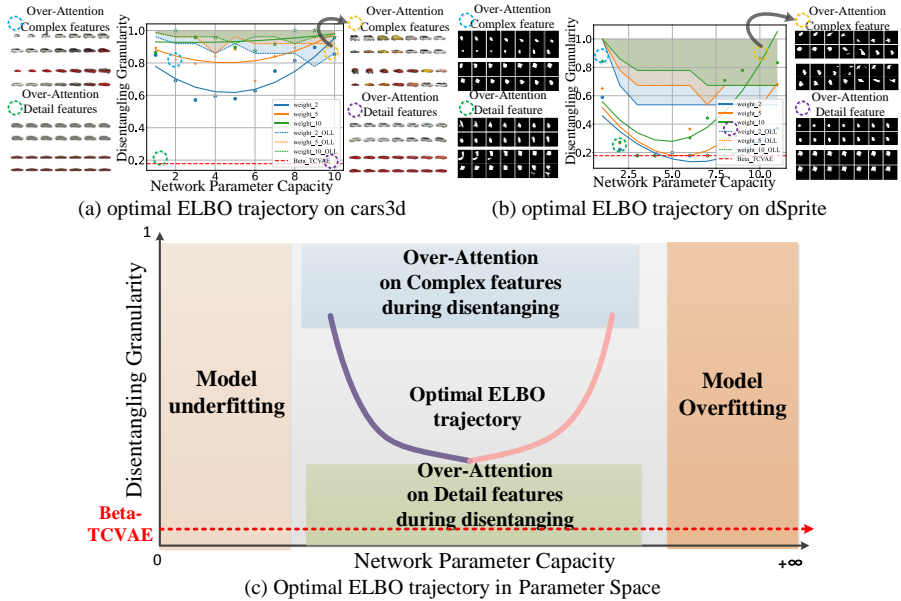

Fig 3. The Fig. 3(a),(b) plot the optimal ELBO trajectory of the model in parameter space constructed from network parameter capacity and disentangling granularity. The green, orange, and blue curves respectively represent the optimal ELBO-fitted curves when the strength weight of the hyperparameter β is equal to 10, 5, and 2. The red dotted line below the figure represents β-TCVAE. The left and right of Fig. 3(a),(b) are images generated by two identical input samples traversing latent variables in four different parameter regions (sampling in the circular dotted area). The Fig. 3(c) shows a general position of the optimal ELBO trajectory and indicates the approximate generation regions for typical samples. In dSprites, cars3d, and shape3d, we marked the occurrence region (colored area in the Fig. 3(a) and Fig. 3(b)) of omniscient latent variable, when $\epsilon$=0.001 and $\delta$=0.01.The green, orange, and blue colored areas represent the areas where omniscient latent variable occurs when the hyperparameter β is equal to 10, 5, and 2 respectively.

## 4. Experiments

### 4.1. Purposes and Basic Settings

We first investigate the impact of varying disentangling granularities on evidence lower bound inference and discuss an intriguing phenomenon discovered while exploring optimal ELBO in parameter space. We then evaluate disentanglement performance using multi-metrics, comparing *β*-STCVAE against several key baselines.

Our experiments are performed on 6 datasets: dSprites [17], celebA [20], shapes3d [19], cars3d [32] ,mnist [33], and mpi3d [33], whose neural network model parameter settings are presented in the appendix as well.

*4.2. Optimal ELBO Trajectory in Parameter Space*

We conducted experiments to observe the degree of disentangling granularity that corresponds to the optimal ELBO for various network parameter capacities, with the unified experimental settings on five datasets as follows:

- {6, 8, 10, 12, 14, 16, 18, 20} is set as the selection list of latent variable number.
- {2, 4, 6, 8, 10} is used to represent the gradually increasing network parameter capacity. For details, please refer to the appendix B.1 Table4.
- 20000-iteration is for a single training round.
- All results are reported by taking the average over 20 trials to avoid bias.

We traversed the parameter capacity selection list and observe the performance of $β$-STCVAE composed of each disentangling granularity. Recording the average disentangling granularity of the model with optimal ELBO under each network parameter capacity, and leveraging a binary function curve to fit these points. It should be reminded that while the disentangling granularity is equal to 1/m (m is the number of latent variables), $β$-STCVAE degenerates into $β$-TCVAE, thus, the training of $β$-TCVAE is included in the experiments.

We trained more than 100K instances of the model across five datasets (detailed results for MNIST and Shape3D could be found in the appendix). Intriguingly, a V-shaped pattern emerged consistently across all datasets. Fig. 3 shows that the new model with a reasonable disentangling granularity could always obtain better ELBO than $β$-TCVAE (red dotted line), regardless of disentangling strength anparameter capacities, except for some situations in the dSprites dataset. It should be noted that the disentangling granularity $g = 1/m$ corresponding to $β$−TCVAE with different latent variable number are inconsistent, so we take the experimental average value 0.0892 as the drawing reference (The vertical axis value of the red dotted line below all the sub-figs in Figure 3).

*4.3. Why Optimal ELBO Trajectory Is V-shaped?*

Intuitively, while the capacity parameter is low, it becomes challenging to capture a fully representative feature within a single latent variable. In this scenario, the model with optimal ELBO tends to relax marginal independent constraints to learn more comprehensive features. The increase in parameter capacity makes the feature capacity of a single latent variable gradually increase, so an objectively complete feature no longer needs a joint distribution composed of redundant latent variables to accommodate. This explains why the disentangling granularity of the optimal ELBO model decreases as parameter capacity gradually increases (purple line area of Fig. 3(c)).

Upon reaching a critical point, the features learned by a single latent variable become sufficiently complex, requiring a correspondingly complex posterior distribution $q(z|x)$ to contain the features. Consequently, relaxing the independence constraint among the one-dimensional Gaussian marginal distributions to form a more complex joint distribution causes the disentangling granularity to rise again (pink line area of Fig. 3(c)). Additionally, we observe that as the disentangling strength $\beta$ enhances, the optimal ELBO trajectory shifts upward. This phenomenon implies that the model allocates more parameter capacity to learn objectively complete features as a balancing mechanism to counteract the high disentangling strength.

In conclusion, we suggest larger disentangling granularity should be selected in the following three situations:

(a). while the total parameter capacity is relatively low. In order to learn more complete data features, the model should tune a larger disentangling granularity;

(b). while the total parameter capacity is relatively high. In order to match and accommodate the complex features that the neural network can learn, a larger disentangling granularity should be selected;

(c). while the disentangling strength is great. A larger disentangling granularity should be selected.

*4.4. Disentangling Expression in Different Regions*

We analyzed the disentangling effect of model sampling within specific regions of the parameter space to understand the impact of disentangling granularity on the model's focus under varying network parameter capacities.

**The blue sampling circle**: model disentangle with high disentangling granularity under low parameter capacity.

**The green sampling circle**: model disentangle with low disentangling granularity under low parameter capacity.

**The yellow sampling circle**: model disentangle with high disentangling granularity under high parameter capacity.

**The purple sampling circle**: model disentangle with low disentangling granularity under high parameter capacity.

After generating numerous samples across various datasets, we observed that models in the upper region of the parameter space tend to disentangle high-complexity features, while those in the lower region tend to disentangle low-complexity features.

Although efficiently allocating parameter capacity remains challenging, we can manipulate disentangling granularity by referring to the regional characteristics of the generated samples. Assuming that the disentangling expression varies smoothly across the entire parameter space, these four specific regions offer valuable insights into the overall trends of the model's disentangling behavior, making it highly interpretable.

In contrast, $\beta$-TCVAE collapses into a straight line within this space due to its inability to adjust disentangling granularity (as indicated by the red dotted line in Fig. 3(a),(b),(c)). This demonstrates that constraining the independence of factorized marginal distributions is insufficient for disentangling complex features.

*4.5. Omniscient Latent Variable*

In the experiments, we also observed that while the latent variable grouping factor reaches a certain threshold, and the parameter capacity is sufficiently high, data features can overload a single latent variable. This variable then exhibits marginal and conditional entropy values close to 0, meaning it provides no useful information. We refer to this phenomenon as the 'omniscient latent variable', where the variable attempts to encapsulate the characteristic states of all samples.

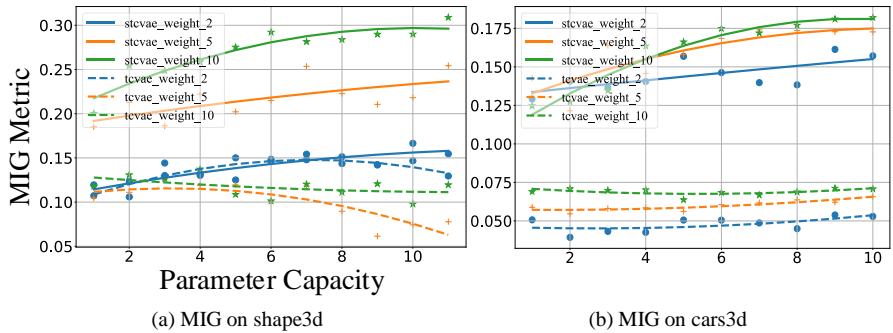

(a) MIG on shape3d    (b) MIG on cars3d

Fig 4. The left, right are the average MIG of the two models under different parameter capacities on shape3d, cars3d datasets. The solid line represents the new model, the dotted line represents β-TCVAE. Green, orange, and blue represent the model with the disentangling strength β is 10, 5, and 2 respectively.

So, we try to locate the occurrence range of this phenomenon in cars3D, dSprites, shape3D datasets with ground true factor labels, and defined the omniscient latent variable's information entropy as satisfying the below probability:

$$P\big[E_{q(z_i)}(-log q(z_i)) < \epsilon\big] \geq 1 - \delta, \tag{8}$$

in all trained models. Fig. 3(b) illustrated that the omniscient latent variable appears in the upper-right region for the dSprites dataset, whereas it was more concentrated toward the top in the other two datasets (see supplementary materials C for experiments on Shape3D). This difference arises from the network architectures used: the dSprites model was trained with an MLP, while the other two datasets were trained with a convolutional network (Conv). Since Conv networks generally had a higher overall parameter capacity than MLPs, this results in a global rightward shift of the parameter space. The presence of the omniscient latent variable in the upper-right region suggested that this phenomenon occurs in areas of over-attention to high-complexity features, aligning with our expectation of disentangling behavior in this region.

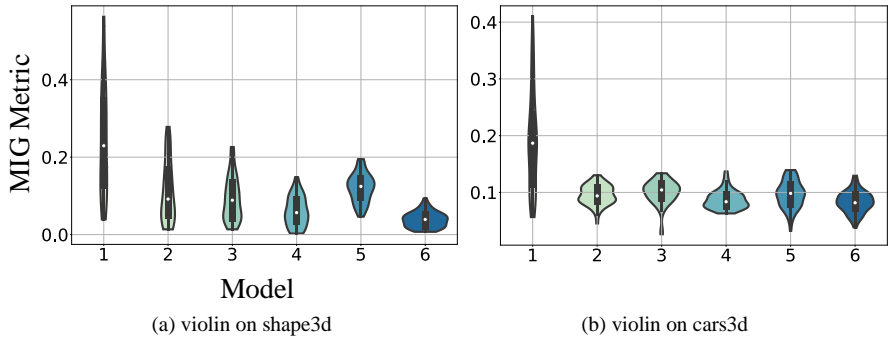

(a) violin on shape3d  (b) violin on cars3d

Fig 5. The Fig. 5(a),(b) are the MIG-based disentangling metric violin graphs of different VAE models on shape3d, cars3d, datasets. Models are abbreviated (1=β-STCVAE_8_2; 2=β-TCVAE_16; 3=β-TCVAE_8; 4=βVAE_8; 5=FactorVAE_8; 6=DIP-I_8). Abbreviated writing method: (Model name)_(latent variables)_(Grouping factor). For example, Stcvae_8_2 represents a Stcvae model with 8 latent variables when the total number of latent dimensions is 16, and the grouping factor is 2. β-TCVAE_8 represents a β-TCVAE model with 8 latent variables.

*4.6. Disentangling Capability Evaluation*

As shown in Fig. 4, the disentanglement of *β*-STCVAE improved significantly with the increase in networks' parameter capacity. In contrast, the disentanglement of *β*-TCVAE showed only modest improvement. We believe the main reason is that under different parameter capacities, *β*-STCVAE could consistently identify the appropriate disentangling granularity of latent variables to contain the data features learned by the model, whereas *β*-TCVAE cannot, which limited the expressiveness of disentangling.

The MIG disentangling metrics of *β*-STCVAE, *β*-TCVAE [20], *β*VAE [17], FactorVAE [19], and DIP-VAE-I [18] were tested on 3 datasets under a fixed number of latent variables, as shown in Fig. 5. The model's hyperparameters were set as described in B.1 of the appendix. The disentangling performances of *β*-STCVAE on these three datasets were significantly improved over factorized independence-constrained VAEs. It is worth noting that *β*-STCVAE exhibits greater variance in MIG compared to other factorized VAEs. We attributed this to the diverse disentanglement granularity configurations used in the experiment, which highlighted the model's broader disentangling adaptability under the MIG metric.

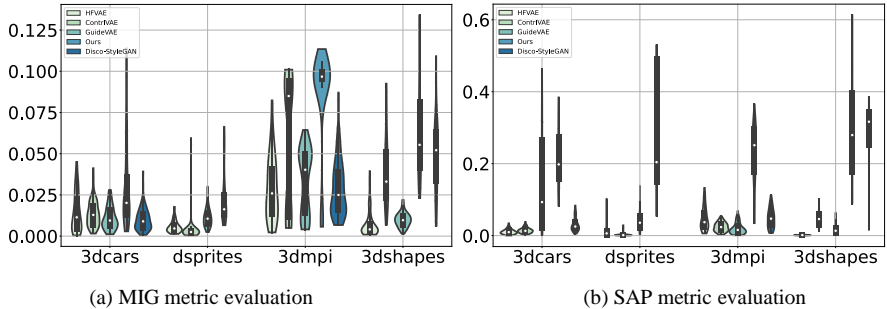

(a) MIG metric evaluation       (b) SAP metric evaluation

Fig 6. We trained five models on four experimental groups (3dcars, dsprites, 3dmpi, 3dshapes), within each experimental groups arranged from left to right as follows: HFVAE, ControlVAE, GuidedVAE (unsupervised), β-STCVAE, and DisCo (based on Style-GAN). The DisCo model experiment is missing from the dsprites dataset because the author does not provide a pre-trained model on Style-GAN for dsprites. We fixed the latent variable dimension of all models to be 8 (the latent variable dimensions of β-STCVAE and HFVAE are 16, the grouping factors of β-STCVAE are 2, and the sub-group size of HFVAE is 2).

This demonstrated that incorporating adjustable disentanglement granularity enhanced the model's flexibility in capturing varying feature complexities.

We also compared the disentangling ability of $\beta$-STCVAE with several recent important baselines: Guide-VAE (Ding et al. 2020), ControlVAE [34], DisCo [35], and HFVAE [36]. The comparison was made using commonly recognized unsupervised disentanglement evaluation metrics, such as MIG [20], Factor [19], and SAP [18], under multiple datasets. The final results are shown in Fig. 6 (each experiment was repeated 100 times with random initialization seeds). The average score and variance of $\beta$-STCVAE across various disentangling metrics were higher than those of other disentanglement baselines with explicit structural priors. This suggested that, even without introducing complex explicit structural prior biases, merely adjusting the disentangling granularity could significantly enhance the disentanglement capabilities of VAEs.

## 5. Conclusion

We uncovered disentangling granularity as a crucial implicit inductive bias in factorized VAEs, governing the model's tendency to disentangle features of varying complexity. Extensive experiments highlighted the impact of disentangling granularity on both disentanglement performance and evidence lower bound inference. Furthermore, we discovered a "V"-shaped optimal ELBO trajectory, which revealed that conventional factorized VAEs induce low-complexity disentanglement due to fixed disentangling granularity, limiting their ability to capture more intricate feature pattern. These findings provided new insights into the role of implicit inductive biases in VAEs, offering a more principled understanding of unsupervised disentanglement.

To date, most studies have focused on making VAEs' disentangling capabilities more interpretable to humans. However, even among humans, the perception of feature disentanglement varies depending on subjective conceptual frameworks. Rather than solely optimizing for predefined evaluation criteria, we advocate for greater controllability in unsupervised disentanglement. We further argue that disentangled representation learning should encourage diversity in unsupervised settings while leveraging explicit supervision for customized disentanglement objectives.

[6] ... *Speech recognition using deep neural networks: A systematic review*, IEEE access, 7 (2019), pp. 19143-19165.
[7] K. XU, L. XU, G. HE, W. YU and Y. LI, *Beyond Alignment: Blind Video Face Restoration via Parsing-Guided Temporal-Coherent Transformer*, arXiv preprint arXiv:2404.13640 (2024).
[8] X.-B. NGUYEN, C. N. DUONG, X. LI, S. GAUCH, H.-S. SEO and K. LUU, *Micron-bert: Bert-based facial micro-expression recognition*, Proceedings of the IEEE/CVF Conference on Computer Vision and Pattern Recognition, 2023, pp. 1482-1492.
[9] M. T. H. FUAD, A. A. FIME, D. SIKDER, M. A. R. IFTEE, J. RABBI, M. S. AL-RAKHAMI, A. GUMAEI, O. SEN, M. FUAD and M. N. ISLAM, *Recent advances in deep learning techniques for face recognition*, IEEE Access, 9 (2021), pp. 99112-99142.
[10] L. LI, X. MU, S. LI and H. PENG, *A review of face recognition technology*, IEEE access, 8 (2020), pp. 139110-139120.
[11] J. LI, J. ZHANG, J. LI, G. LI, S. LIU, L. LIN and G. LI, *Learning background prompts to discover implicit knowledge for open vocabulary object detection*, Proceedings of the IEEE/CVF Conference on Computer Vision and Pattern Recognition, 2024, pp. 16678-16687.
[12] X. LIU, C. ZHENG, M. QIAN, N. XUE, C. CHEN, Z. ZHANG, C. LI and T. WU, *Multi-View Attentive Contextualization for Multi-View 3D Object Detection*, Proceedings of the IEEE/CVF Conference on Computer Vision and Pattern Recognition, 2024, pp. 16688-16698.
[13] S. ZHANG, *Current status and development trend of target tracking methods based on deep learning*, International Conference on Computational Modeling, Simulation, and Data Analysis (CMSDA 2021), SPIE, 2022, pp. 489-493.
[14] R. TANG, L. LIU, A. PANDEY, Z. JIANG, G. YANG, K. KUMAR, P. STENETORP, J. LIN and F. TURE, *What the daam: Interpreting stable diffusion using cross attention*, arXiv preprint arXiv:2210.04885 (2022).
[15] Z. XU, *Research on deep learning in natural language processing*, Advances in Computer and Communication, 4 (2023).
[16] E. ARKHANGELSKAYA and S. I. NIKOLENKO, *Deep learning for natural language processing: a survey*, Journal of Mathematical Sciences, 273 (2023), pp. 533-582.
[17] I. HIGGINS, L. MATTHEY, A. PAL, C. P. BURGESS, X. GLOROT, M. M. BOTVINICK, S. MOHAMED and A. LERCHNER, *beta-vae: Learning basic visual concepts with a constrained variational framework*, ICLR (Poster), 3 (2017).
[18] A. KUMAR, P. SATTIGERI and A. BALAKRISHNAN, *Variational inference of disentangled latent concepts from unlabeled observations*, arXiv preprint arXiv:1711.00848 (2017).
[19] H. KIM and A. MNIH, *Disentangling by factorising*, International conference on machine learning, PMLR, 2018, pp. 2649-2658.
[20] R. T. CHEN, X. LI, R. B. GROSSE and D. K. DUVENAUD, *Isolating sources of disentanglement in variational autoencoders*, Advances in neural information processing systems, 31 (2018).

**Appendix**

Here we provide more details about our paper. An overview of this supplementary material is as follows:

**Section A**: Total Correlation Decomposition Details
    -A.1 Total Correlation Bottom-Up Path Decomposition
**Section B**: Experimental Settings
    -B.1 Experimental Model Design Details
**Section C**: Supplementary Experiments
    -C.1 V-Shaped optimal ELBO on Shape3D
    -C.2 V-Shaped optimal ELBO on Mnist
    -C.3 Supplementary experiments under disentangling metrics

**A Total Correlation Decomposition Details**

**A.1 Total Correlation Bottom-Up Path Decomposition**

Suppose the total number of $z$ is $n$. $p(z_{-i,j})$ represents the joint distributions of all variables except $z_i, z_j$.

$$D_{KL}(q(z) || \prod_k q(z_k)) = \int_{-\infty}^{+\infty} q(z) log \frac{q(z)}{\prod_k q(z_k)}. \tag{1}$$

Objective (1) is an improper integral with upper and lower bounds $(-\infty, +\infty)$. Since $q(z)$ has no analytical formula, the solution σ of $D_{KL}(q(z) || \prod_k q(z_k))$ can only be calculated using the sampling method.

While σ is a constant, it means that objective (1) converges, then the following derivations:

$$D_{KL}(q(z)|| \prod_k q(z_k)) = \int_{-\infty}^{+\infty} q(z) log \frac{q(z)}{\prod_k q(z_k)} = \lim_{\substack{u \to +\infty \\ v \to -\infty}} \int_v^u q(z) log \frac{q(z)}{\prod_k q(z_k)} = \sigma,$$

(2)

where $\forall \varepsilon > 0$, $\exists \alpha > 0$, so that for $\forall u > \alpha$, and $\forall v < -\alpha$, there is $|\int_v^u q(z) log \frac{q(z)}{\prod_k q(z_k)} - \sigma| < \varepsilon$. Therefore, it can be assumed that the probability density function of each dimension of the latent variable $z_i$ of the latent space distribution $z$ has a unified upper and lower limit $(v, u)$ so that the gap between $\int_v^u q(z) log \frac{q(z)}{\prod_k q(z_k)}$ and $D_{KL}(q(z)|| \prod_k q(z_k))$ is smaller than an extremely small positive number $\varepsilon$.

Therefore, a definite integral with upper and lower bounds can be used to replace the anomalous integral. For the sake of brevity, the integral symbol $\int$ in the following derivations represents the definite integral to integrate all dimensions $(z_1, z_2 \dots z_n)$ of $z$ respectively.

$$\begin{aligned}
D_{KL}(q(z)|| &\prod_k q(z_k)) \\
&= \int q(z) log \frac{q(z)}{\prod_k q(z_k)} \\
&= \int q(z) log \frac{q(z_1, z_2) q(z_{-1,2}|z_1, z_2)}{\prod_k q(z_k)} \\
&= \int q(z) log \frac{q(z_1, z_2)}{q(z_1) q(z_2)} + \boxed{\int q(z) log \frac{q(z)}{\prod_{\substack{k \neq 1 \\ k \neq 2}} q(z_k) \cdot q(z_1, z_2)}},
\end{aligned}$$

(3)

where the box in objective (3) can be decomposed as:

$$\begin{aligned}
&= \int q(z) log \frac{q(z_3, z_4) q(z_{-3,4}|z_3, z_4)}{\prod_{\substack{k \neq 1 \\ k \neq 2}} q(z_k) \cdot q(z_1, z_2)} \\
&= \int q(z) log \frac{q(z_3, z_4)}{q(z_3) q(z_4)} + \boxed{\int q(z) log \frac{q(z_1, z_2 \dots z_n)}{\prod_{\substack{k \neq 1 \\ k \neq 2 \\ k \neq 3 \\ k \neq 4}} q(z_k) \cdot q(z_1, z_2) \cdot q(z_3, z_4)}}.
\end{aligned}$$

(4)

We can continue to decompose the box in objective (4). Finally, we get an objective that needs to be discussed in different situations.

If *n* is even, then:

$$= \int q(z)\log\frac{q(z_1,z_2)}{q(z_1)q(z_2)} + \int q(z)\log\frac{q(z_3,z_4)}{q(z_4)q(z_4)} + \cdots + \boxed{\int q(z)\log\frac{q(z)}{q(z_1,z_2)\cdot q(z_3,z_4)\dots q(z_{n-1},z_n)}}.$$

(5)

If $n$ is odd, then:

$$= \int q(z)\log\frac{q(z_1,z_2)}{q(z_1)q(z_2)} + \int q(z)\log\frac{q(z_3,z_4)}{q(z_3)q(z_4)} + \cdots$$

$$+ \boxed{\int q(z)\log\frac{q(z)}{q(z_1,z_2)\cdot q(z_3,z_4)\dots q(z_{n-2},z_{n-1})\cdot q(z_n)}}$$

(6)

While the box in objective (5) is equal to 0, it means that except for the last item, all the $q(z_i, z_j)$ joint distributions that appear in the previous items are independent. In this situation, the underlined parts in objective (5) can be deduced:

$$\int q(z)\log\frac{q(z_i,z_j)}{q(z_i)q(z_j)} = E_{q(z)}[\log\frac{q(z_i,z_j)}{q(z_i)q(z_j)}]$$

$$= \int q(z_i,z_j)\log\frac{q(z_i,z_j)}{q(z_i)q(z_j)}q(z_{-i,j}) = \int q(z_i,z_j)\log\frac{q(z_i,z_j)}{q(z_i)q(z_j)} = I(z_i\,;\,z_j) =$$

$$E_{q(z_i,z_j)}\left[\log\frac{q(z_i,z_j)}{q(z_i)q(z_j)}\right],$$

(7)

where $i \neq j$, $i$ and $j < n$.

While the box in objective (5) is not equal to 0, it means that except for the last item, the joint distributions $q(z_i, z_j)$ appearing in the previous items are not independent. The underlined parts in objective (5) can be deduced:

$$\int q(z)\log\frac{q(z_i,z_j)}{q(z_i)q(z_j)} = E_{q(z)}[\log\frac{q(z_i,z_j)}{q(z_i)q(z_j)}]$$

$$= E_{q(z_i,z_j)}[q(z_{-i,j}|z_i,z_j)\cdot\log\frac{q(z_i,z_j)}{q(z_i)q(z_j)}].$$

(8)

Intuitively, taking $q(z_{-i,j}|z_i,z_j)$ as the importance weight. The importance weight is 1, while $q(z_{-i,j})$ and $q(z_i,z_j)$ are independent of each other. As a result, while $q(z_i,z_j)$ and $q(z_{-i,j})$ are not independent, it can be viewed as a way of describing $I(z_i\,;\,z_j)$ by importance sampling. The derivations of odd case objective (6) is similar to even case objective (5).

Then, we can simplify the above derivations to objective (9). $z_i$ represents the $i_{th}$ dimension's latent variable in the multidimensional latent variable $z$. $cp(i,j)$ represents a mechanism for selecting a pair of latent variables $z_i, z_j$ without replacement, and the $cp(i,j)$ in the summation symbol is consistent with the result of the $cp(i,j)$ in the cumulative multiplication symbol; $z_r$ represents the final remaining latent variable $r$ after selecting several times without replacement under the odd case $cp(i,j)$:

$$TC(z) = D_{KL}(q(z) || \prod_k q(z_k))$$

$$= \begin{cases} \sum_{cp(i,j)} E_{q(z)}[\log \frac{q(z_i, z_j)}{q(z_i)q(z_j)}] + D_{KL}(q(z) || \prod_{cp(i,j)} q(z_i, z_j)) & , n \text{ is even} \\ \sum_{cp(i,j)} E_{q(z)}[\log \frac{q(z_i, z_j)}{q(z_i)q(z_j)}] + D_{KL}(q(z) || \prod_{cp(i,j)} q(z_i, z_j) \cdot q(z_r)) & , n \text{ is odd} \end{cases},$$

(9)

where $E_{q(z)}[\log \frac{q(z_i, z_j)}{q(z_i)q(z_j)}]$ can be regarded as calculating $I(z_i ; z_j)$ using importance sampling, thus:

$$TC(z) = \begin{cases} \sum_{cp(i,j)} I(z_i ; z_j)|_{q(z) = q(z_i, z_j) \cdot q(z_{-i,j})} + D_{KL}(q(z) || \prod_{cp(i,j)} q(z_i, z_j)) & , n \text{ is even} \\ \sum_{cp(i,j)} I(z_i ; z_j)|_{q(z) = q(z_i, z_j) \cdot q(z_{-i,j})} + D_{KL}(q(z) || \prod_{cp(i,j)} q(z_i, z_j) \cdot q(z_r)) & , n \text{ is odd} \end{cases},$$

(10)

where the $q(z_i, z_j)$ and $q(z_{-i,j})$ in the target distribution are independent, but they are not necessarily independent in the proposed distribution. $I(z_i ; z_j)|_{q(z) = q(z_i, z_j) \cdot q(z_{-i,j})}$ indicates that the importance sampling method is used to calculate the $I(z_i ; z_j)$ that satisfy the target distribution under the proposed distribution $q(z_i, z_j)$.

Then set $TC_{joint\_1}(z)$ as:

$$TC_{joint\_1}(z) = \begin{cases} D_{KL}(q(z) || \prod_{cp(i,j)} q(z_i, z_j)) & , n \text{ is even} \\ D_{KL}(q(z) || \prod_{cp(i,j)} q(z_i, z_j) \cdot q(z_r)) & , n \text{ is odd} \end{cases}. \quad (11)$$

We replace $\sum_{cp(i,j)} I(z_i ; z_j)|_{q(z) = q(z_i, z_j) \cdot q(z_{-i,j})}$ with $MU_{joint\_1}(z)$, then $TC_{joint\_1}(z) = TC(z) - MU_{joint\_1}(z)$. $MU_{joint\_1}(z)$ is always positive, defined as the accumulation of mutual information. Thus, the total correlation of the joint distribution must be less than or equal to the total correlation of the marginal distribution, which can be represented as $TC_{joint\_1}(z) \leq TC(z)$.

The decomposition has an iterative relationship and can continue by treating $TC_{joint\_1}(z)$ as a new $TC(z)$. All joint distributions $q(z_i, z_j)$ selected by $cp(i,j)$ in the $TC_{joint\_1}(z)$ are regarded as the elements of new latent variables' distributions set $\{q(z_{1\_1}), q(z_{1\_2})..q(z_{1\_n/2})\}$ ($(n+1)/2$ variables in total while $n$ is odd, and $n/2$ in total while $n$ is even), thus we have:

$$TC_{joint\_2}(z) = TC_{joint\_1}(z) - MU_{joint\_2}(z),$$
$$TC_{joint\_3}(z) = TC_{joint\_2}(z) - MU_{joint\_3}(z),$$
$$.... \tag{12}$$

finally, there is the following objective:

$$TC(z) = MU_{joint\_1}(z) + MU_{joint\_2}(z) + ... + I(z_{f\_1}; z_{f\_2}), \tag{13}$$

where $f = int(log_2 n) - 1$. If $f < 1$, then $z_{f\_i} = z_i$, and $n$ is the total dimension of $z$, and $int(c)$ is the rounded-up integer of $c$.

Only two variables will be left in the final round of decomposition regardless of the parity of $n$, represented by $z_{f\_1}$ and $z_{f\_2}$ in objective (13). $q(z_{f\_1})$ and $q(z_{f\_2})$ are two joint distributions composed of distributions selected by the $cp(i,j)$ of the $f$th round. Notice that, only the mutual information $I(z_{f\_1}; z_{f\_2})$ obtained in the last round of decomposition is not based on importance sampling. All $MU_{joint}(z)$ are based on importance sampling.

## B Experimental Settings

### B.1 Experimental Model Design Details

The main experiment's neural network models are divided into two types: MLP model and Conv model. dSprites, mnist, and celebA are trained with MLP neural network, cars3d, shape3d are trained with Conv neural networks. The parameter capacities in the encoder and decoder are variable to make it easier to observe the impact of changes in model parameter capacity. In Table1 and Table2 "Layer" represents the number of hidden layers in model; "neuron_num" represents the number of neurons in a single layer, also the value in the "parameter capacity list" in the main body and Table 4. The neuron_num and layer variables determine the parameter capacity of the model.

| Encoder | Decoder |
|---|---|
| Input: width * hight * channel | Input: *Latent_dim* |
| FC *neuron_num*,ReLU | FC *neuron_num*,Tanh |
| FC *neuron_num*,ReLU * *layer* | FC *neuron_num*,Tanh * *layer* |
| FC *Latent_dim* *2 | FC width * hight * channel |
|  | Reshape width * hight * channel |

*neuron_num, Latent_dim, layer* are Parameters to be preset
* *layer* Represent the number of repetitions of this layer

Table 1. MLP Model

| Encoder | Decoder |
|---|---|
| Input: 64*64* channel | Input: number of *Latent_dim* |
| 4*4 conv,16 LeakyReLU,stride 2 | FC *neuron_num* ,Tanh * *layer* |
| 4*4 conv,32 LeakyReLU,stride 2 | FC 8* 8 * 64 ,Tanh |
| 4*4 conv,64 LeakyReLU,stride 2 | 4*4 upconv,32 |
| FC *neuron_num*,ReLU * *layer* | LeakyReLU,stride 2 |
| FC number of *Latent_dim* *2 | 4*4 upconv,16 |
| | LeakyReLU,stride 2 |
| | 4*4 upconv,channel LeakyReLU,stride 2 |

*neuron_num, Latent_dim, layer* are Parameters to be preset
* *layer* Represent the number of repetitions of this layer

Table 2. Conv Model

| Model | Parameter | Value |
|---|---|---|
| $\beta-STCVAE, \beta-TCVAE, \beta VAE$ | $\beta$ | [1,2,4,6,8,10] |
| FactorVAE | $\Gamma$ | [10,20,40,60,80,100] |
| DIP-VAE-I | $\lambda od$ | [1,2,5,10,20,50] |
| | $\lambda d$ | $10\lambda od$ |
| ContrlVAE | $\beta_{min}$ | [1,2,5] |
| GuidedVAE | $d_{zdef}, d_{zcont}$ | $d_{zdef} = d_{zcont} = 4$ |
| DisCo | $\tau, \lambda, Led$ | 1 |
| | $T$ | 0.95 |
| | $N, K, D$ | 64 |

Table 3. Hyperparameter settings of different models

| dataset | model | Hidden layer | Parameter capacity list |
|---|---|---|---|
| mnist | MLP | 1layer | [20,40,60,80,100,200,300,400,500,1000,1500, 2000,2500,3000,3500,4000,4500,5000] |
| celebA | MLP | 1layer | [500,1000,1500,2000,2500,3000,3500,4000,4500, 5000] |
| dSprite | MLP | 1layer | [100,500,1000,1500,2000,2500,3000,3500,4000, 4500,5000] |
| Cars3D | Conv | 2layer | [100,200,300,400,500,600,700,800,900,1000] |
| shape3D | Conv | 2layer | [200,300,400,500,600,700,800,900,1000,1100,1200] |

Table 4. Model and parameter capacity selection for different training set

## C Supplementary Experiment

### C.1 V-shaped optimal ELBO on Shape3D

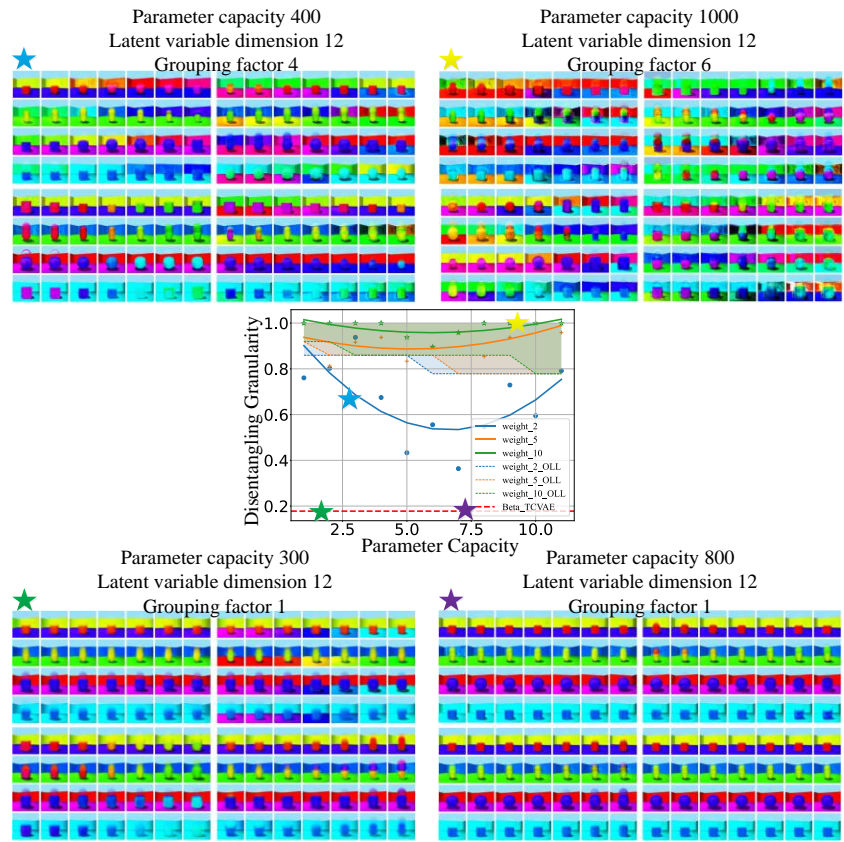

Fig 7. While the disentangling strength is 2 on the shape3d dataset. The parameter training model for 4 cases (blue, yellow, green, purple) is sampled in the parameter space, analyzing the characteristics of their generated samples. Extract 4 input samples, and latent traverse on range [-2, +2] for 4 variables on the model, and all other variables remain 0.

## C.2 V-shaped optimal ELBO on Mnist

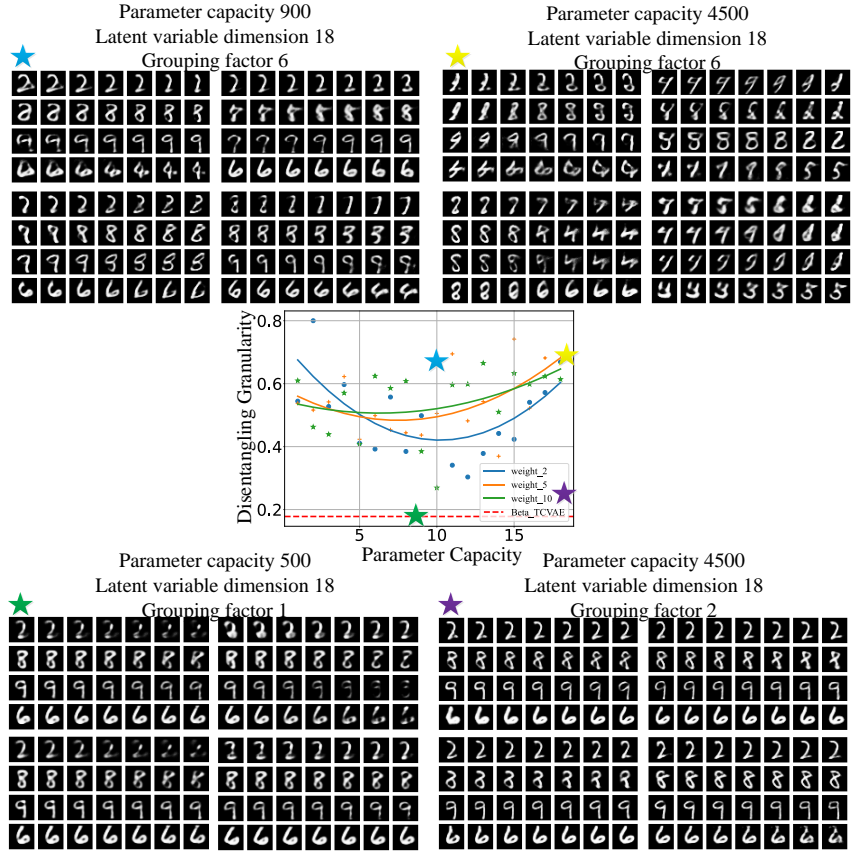

Fig 8. While the disentangling strength is 10 on the mnist dataset. The parameter training model for 4 cases (blue, yellow, green, purple) is sampled in the parameter space, analyzing the characteristics of their generated samples. Extract 4 input samples, and latent traverse on range [-2, +2] for 4 variables on the model, all other variables remain 0. All latent traversing on the blue model has a transform in digital category; however, the lower right purple model latent traversing only exhibits a shift in style detail resembling HFVAE; the green model near the lower left corner generates low-quality fuzzy samples; the yellow model in the upper right corner shows the phenomenon of omniscient latent variable.

## C.3 Supplementary experiments under disentangling metrics

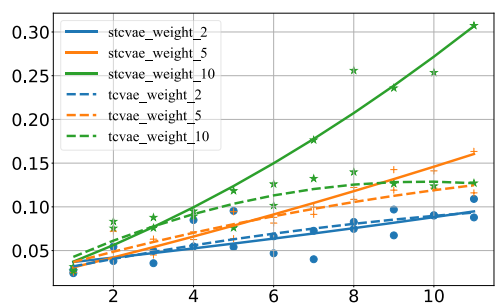

Fig 9. the average MIG of the two models under different parameter capacities on dSprites.

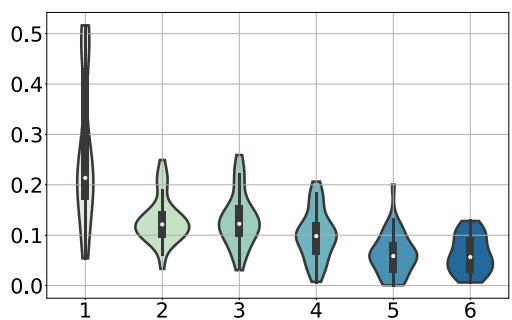

Fig 10. the MIG-based disentangling metric violin graphs on dSprites of different VAEs

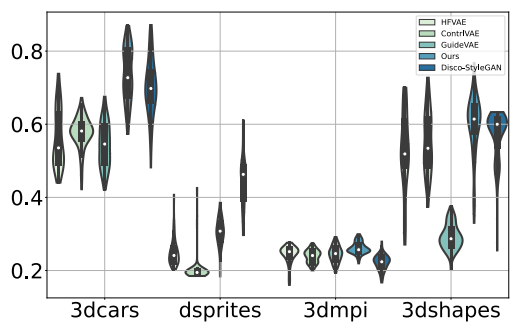

Fig 11. Factor metric evaluation with five models divide in four experimental groups